\documentclass[letterpaper]{article} 
\usepackage{aaai23}  
\usepackage{times}  
\usepackage{helvet}  
\usepackage{courier}  
\usepackage[hyphens]{url}  
\usepackage{graphicx} 
\usepackage{natbib}  
\usepackage{caption} 
\usepackage{algorithm}
\usepackage{algorithmic}

\usepackage{subfig}
\usepackage{tabularx}

\usepackage{newfloat}
\usepackage{listings}
\DeclareCaptionStyle{ruled}{labelfont=normalfont,labelsep=colon,strut=off} 
\floatstyle{ruled}
\newfloat{listing}{tb}{lst}{}
\floatname{listing}{Listing}
\title{Utilizing Generative Adversarial Networks for Stable Structure Generation in Angry Birds}
\author {
    Frederic Abraham \textsuperscript{\rm 1}, Matthew Stephenson \textsuperscript{\rm 2}
}
\affiliations {
    \textsuperscript{\rm 1} Maastricht University, Department of Advanced Computing Sciences, Maastricht, the Netherlands\\
    \textsuperscript{\rm 2} Flinders University, College of Science and Engineering, Adelaide, Australia\\
    fm.abraham@alumni.maastrichtuniversity.nl, matthew.stephenson@flinders.edu.au
}

\begin{document}

\maketitle

\begin{abstract}
This paper investigates the suitability of using Generative Adversarial Networks (GANs) to generate stable structures for the physics-based puzzle game Angry Birds. While previous applications of GANs for level generation have been mostly limited to tile-based representations, this paper explores their suitability for creating stable structures made from multiple smaller blocks. This includes a detailed encoding/decoding process for converting between Angry Birds level descriptions and a suitable grid-based representation, as well as utilizing state-of-the-art GAN architectures and training methods to produce new structure designs. Our results show that GANs can be successfully applied to generate a varied range of complex and stable Angry Birds structures.
\end{abstract}

\section{Introduction} 

Procedural Content Generation (PCG), which describes the creation of content through algorithmic means, has become an increasingly prominent aspect of video game development \cite{Amato2017}.
In the same timeframe as conventional PCG techniques were being researched, the number of applications utilizing Machine Learning (ML) approaches has also increased, with techniques such as Neural Networks and Deep Learning receiving a large amount of attention \cite{Goodfellow-et-al-2016}.
Consequently, the use of ML approaches for content generation, under the abbreviation PCGML, has become an area of significant research interest \cite{Summerville2017}.

While many different types of ML algorithms have been used to generate content, one of the most promising approaches in recent years has been the use of Generative Ad-
versarial Networks (GANs) \cite{GuiSun2021,Jabbar2020,Saxena2020}.
A GAN is the construct of two adversarial networks, in which one network generates new content and the second network discriminates the generated content to differentiate between real and generated \cite{Goodfellow2014}.
While the primary application of GANs has largely been for image and video synthesis \cite{He2018,article123}, they have also been applied to several other domains including video game content generation \cite{Giacomello2018,Volz2018}. However, to the best of our knowledge, when it comes to generating level-based content for video games, GANs have only been applied to discrete domains with no physical constraints.

In this paper, we present an approach for generating stable structures made of several rectangular blocks in a continuous physical environment. More specifically, we generate structures for use in the 2D physics-based puzzle game Angry Birds \cite{rovio}. This domain provides a unique and novel challenge for applying GAN-based approaches, where considerations such as the game's continuous environment space and physical constraints must be taken into account. Our results demonstrate that such an approach is viable for generating complete Angry Birds structures.

The remainder of this paper is organized as follows. We first describe background details and prior work related to GANs and their use in level generation, as well as previous approaches to Angry Birds level generation. Next, we describe the methodology we employed with regards to level encoding and decoding, as well as our GAN model training process. We then describe our experiments, providing details on how our specific GAN model was trained and analyze the generated levels it was able to produce. We finish with a summary conclusion of our approach and output, along with suggestions for future work.

\section{Background and Related Work} 

\subsection{Generative Adversarial Networks}

A GAN is a framework for producing generative models via a process in which two networks, the generator and the discriminator, are trained simultaneously \cite{Goodfellow2014}. The generator aims to capture the data distribution of the training data, while the discriminator, also called the critic, tries to differentiate between samples drawn from the training data and samples generated by the generator. The generator is trained to maximize the probability that the discriminator mistakes its generated example as drawn from the actual distribution. The central concept behind applying GANs is to define the given task as a game between two opposing
systems, which are then trained in an adversarial manner to reach a zero-sum Nash equilibrium
\cite{Moghadam2021}. While GANs have been primarily used for image synthesis \cite{Goodfellow2014,Radford2015,Liu2016,Karras2017,He2018}, they have also been successfully applied to many other applications including object detection \cite{objectdetection1, POSILOVIC2021361}, natural language processing \cite{subramanian-etal-2017-adversarial}, audio enhancement \cite{audio1,Biswas2020AudioCE}, anomaly detection \cite{anom1,Xia2020}, and, most relevant for this paper, video game level generation.

In recent years, several variations and improvements to GANs have been proposed for different aspects of the underlying framework.
The architecture of each network, the controllability, the training process, scaling, adaptation and application in different domains are a few research directions that have been investigated. Some notable advancements include the development of Deep Convolutional GANs (DCGANs) \cite{Radford2015}, which introduced deeper architectures and improved training stability for image synthesis tasks. Another significant improvement came with the introduction of Wasserstein GANs (WGANs) \cite{Arjovsky2017Wgan}, which proposed a new objective function to address training difficulties and mode collapse issues \cite{Salimans2016}. The term mode collapse refers to the situation where a GAN model repeatedly generates highly similar outputs that don't represent the variety of content present in the original training set.

\subsection{Level Generation with GANs}

While GANs have been used to generate a variety of different types of game content, such as the generation of NPC character sprites \cite{KIM2023118491, Coutinho_Chaimowicz_2022},
we will focus primarily on the use of GANs to generate game levels.
One of the first to apply GANs in
the context of level generation was Giacomello et al.,
in their work to generate DOOM level images based on human-designed examples \cite{Giacomello2018}. They describe their preliminary results as
a good starting point for researching the viability of GANs
compared to classical PCG. The generated level images contained
DOOM typical features and are reportedly interesting to explore, although the generated data could not be decoded
into playable levels.

Volz et al. also utilized GANs to generate complete Mario levels \cite{Volz2018}. They use a
Covariance Matrix Adaptation Evolutionary Strategy (CMA-ES) to search the latent
space of the GANs generator to influence the outcome based on different metrics
over the generated levels. Their first approach is to optimize different block distributions. For example, fewer stone blocks could lead to an air level with greater
difficulty. In their second approach, they utilize a Mario AI \cite{Togelius2013}
that can produce playthrough data of their generated levels. They focused on optimizing toward playable levels with a scalable difficulty. The idea of using latent
variable evolution (LVE) to explore the generator’s latent space was first introduced by Bontrager et al., in their works to match generated fingerprints to
as many real fingerprints as possible \cite{Bontrager2017}. Evolving the latent space to gain control over
the output stands in contrast to Conditional GANs (CGANs) \cite{Mehdi2014}, which utilize a condition vector combined with the noise vector as input to
the generator to produce a controllable output.
They conclude that GANs can capture high-level structures of the training level, although they may sometimes produce broken elements such as incomplete pipes and
structures.

\begin{figure}
  \includegraphics[width=1.0\linewidth]{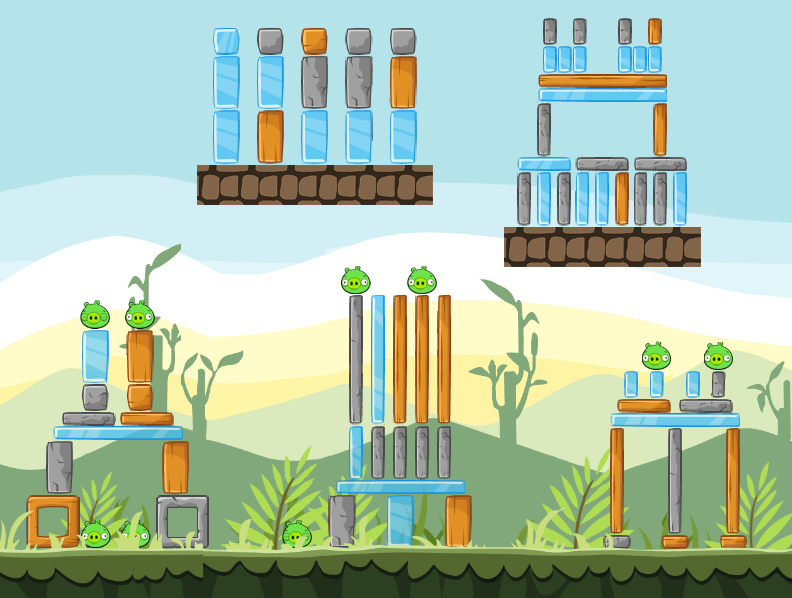}
  \caption{An example Angry Birds level containing five distinct structures.}
  \label{fig:AB_example_level}
\end{figure}

\begin{table}
\begin{tabular}{|p{3cm}|p{2.7cm}|p{1.8cm}|p{1.8cm}|}
\hline
\multicolumn{1}{|l|}{\textbf{Id}} & \textbf{Shape} & \textbf{Name} & \textbf{Dimensions}    \\  \hline 
\multicolumn{1}{|l|}{1}  & \parbox[c]{\hsize}{\includegraphics[width=0.05\textwidth]{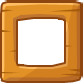}} & SquareHole & (0.85, 0.85) \\  \hline
\multicolumn{1}{|l|}{2}  & \parbox[c]{\hsize}{\includegraphics[width=0.15\textwidth]{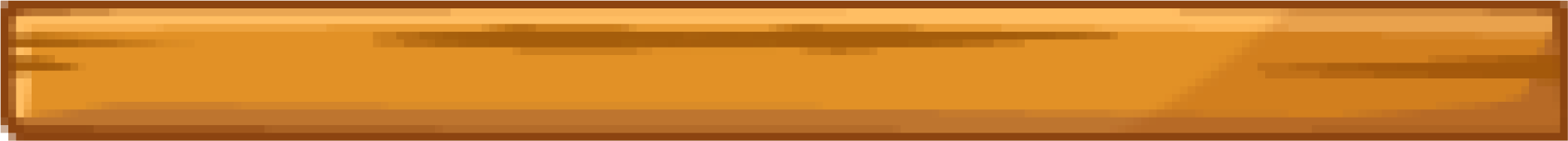}} & RectBig & (2.06, 0.22)    \\  \hline
\multicolumn{1}{|l|}{3}  & \parbox[c]{\hsize}{\includegraphics[width=0.125\textwidth]{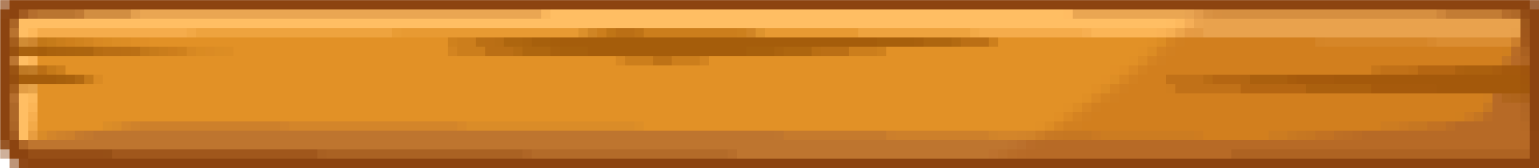}} & RectMedium & (1.68, 0.22)  \\ \hline
\multicolumn{1}{|l|}{4}  & \parbox[c]{\hsize}{\includegraphics[width=0.08\textwidth]{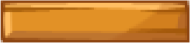}} & RectSmall  & (0.85, 0.2)  \\ \hline
\multicolumn{1}{|l|}{5}  & \parbox[c]{\hsize}{\includegraphics[width=0.075\textwidth]{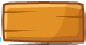}} & RectFat   & (0.85, 0.43)   \\ \hline
\multicolumn{1}{|l|}{6}  & \parbox[c]{\hsize}{\includegraphics[width=0.04\textwidth]{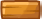}} & RectTiny  & (0.42, 0.22)    \\ \hline
\multicolumn{1}{|l|}{7}  & \parbox[c]{\hsize}{\includegraphics[width=0.02\textwidth]{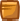}} & SquareTiny & (0.22, 0.22)  \\ \hline
\multicolumn{1}{|l|}{8}  & \parbox[c]{\hsize}{\includegraphics[width=0.04\textwidth]{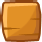}} & SquareSmall & (0.43, 0.43) \\ \hline
\end{tabular}
\caption{Block types that are available in Science Birds.}
\label{fig:block_types}
\end{table}

\subsection{Angry Birds Level Generation}
Angry Birds is a 2D physics-based simulation game, where players are tasked with shooting birds at structures made of multiple smaller blocks. These blocks are each made of a specific material (wood, ice or stone) which affects their resistance to certain bird types. The players objective in each level is to kill all of the pigs, which are often located within or on top of structures, using a limited number of birds. An example Angry Birds level, containing five structures and nine pigs, is shown in Figure \ref{fig:AB_example_level}. Due to the fact that the original Angry Birds game is not open source, most research on Angry Birds instead uses a Unity-based clone of the game called Science Birds, created by Lucas Ferreira \cite{ferreira_2014_a}. This clone contains all the same elements as the original Angry Birds game, and each level is described in a single xml file containing the position of each object.

Table \ref{fig:block_types} shows the available block types that structures can be made from, along with their names and dimensions (width, height). While there are several additional ``irregular'' block shapes that are also available in Angry Birds, specifically two triangular and two circular block types, the vast majority of Angry Birds structures do not include these. Many prior structure generators for Angry Birds either excluded irregular blocks entirely or used them in a purely decorative manner, as the stability of a structure is much harder to verify if irregular blocks are present. In addition to this, blocks are almost always placed at 90-degree angles, once again to reduce the risk of structural instability. As a result of this, our proposed GAN-based generator will be solely trained on, and will therefore only output, structures made of the block types specified in Table \ref{fig:block_types}, placed in either a horizontal (angle = 0) or vertical (angle = 90) orientation.

Over the past decade a large number of different level generators for Angry Birds have been proposed, with many of them being entered into the AIBirds Level generation competition \cite{Stephenson2019}. These generators have experimented with a variety of different approaches, including Genetic Algorithms \cite{ferreira_2014_a}, search-based techniques \cite{Stephenson2016-1,Stephenson2016-2,Stephenson2017}, Monte Carlo Tree Search \cite{Caramanis2016ProceduralCG}, and latent variable evolution in a variational auto encoder \cite{Tanabe2021}, as well as focusing on a variety of different desirable aspects, such as levels that contain structures which resemble quotes or formulas \cite{8080429}, deceptive elements \cite{COG9619031}, or Rube Goldberg Machine mechanisms \cite{8847996}.

\section{Methodology}

This section describes the different components that make up our proposed GAN framework for Angry Birds levels generation. This includes level encoding, level decoding and GAN Model Training.

\subsection{Level Encoding}

This section describes the encoding process that was used to convert an XML level description of a structure into a suitable grid-based data representation that can be used to train our proposed GAN model.

\subsubsection{Grid-Based Representation}

\begin{figure}
  \includegraphics[width=1.0\linewidth]{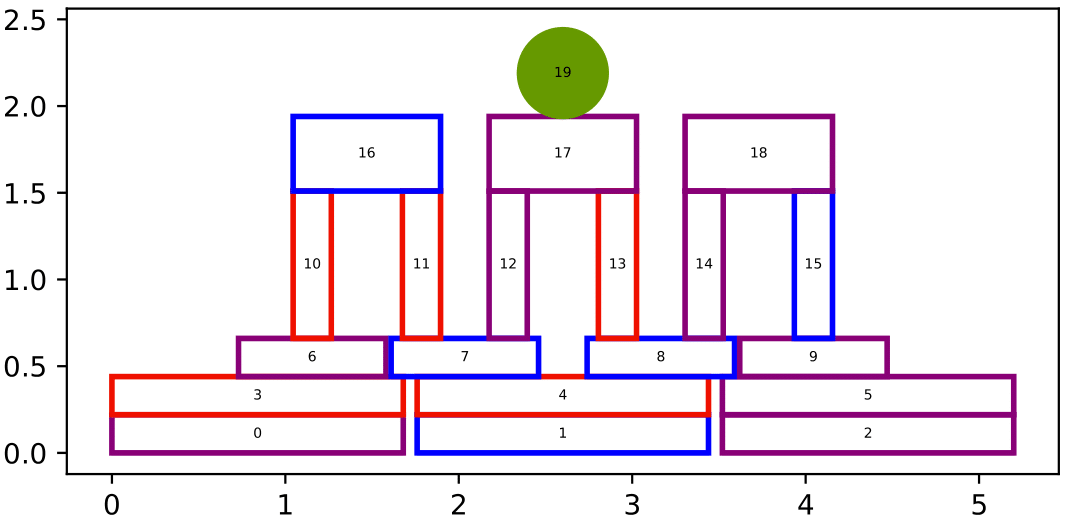}
  \caption{An example Angry Birds structure represented as a colored wireframe diagram.}
  \label{fig:wireframe}
\end{figure}

One of the major challenges in this domain is the real-valued positioning and dimensions of in-game objects. Continuous values such as these are impractical for GAN models, which typically require the input to be in a discrete, grid-based representation to function effectively (e.g., the pixels of an image, or the individual tiles in a Mario level). As a result, our XML structure representation first needs to be converted or encoded to a grid-based representation with a specified degree of precision (i.e., a raster size). Figure \ref{fig:wireframe} shows an example of a simple Angry Birds structure with continuous block positions and dimensions, represented visually as a wireframe diagram with different colors for each block material and pigs, that will be used as a reference throughout this section.

In order to discretize a structure to a specified level of precision, a suitable raster size has to be chosen. A smaller raster size better represents the blocks true dimensions, and is able to capture small gaps and imperfect positionings that a larger raster size may miss. The downside of a smaller raster size is that it can dramatically increase the dimensionality of the encoded structure representation, requiring a larger GAN model that results in significantly longer training and generation times. Conversely, a bigger raster size results in a smaller encoded output representation where only a few grid cells represent any given block. This comes with a loss of finer details, that may potentially result in incorrectly represented block sizes.

As a suitable compromise between these two considerations, we selected the largest raster size that resulted in a near-integer value when dividing by any block’s true dimension. This was done to reduce the risk of the same block type being encoded at different sizes, depending on its relative position to the grid lines. For the block sizes present in ScienceBirds a raster size 0.07 resulted in quotients with only a small remainder, and was subsequently chosen as our raster size. This raster size of 0.07 in-game distance units, can be equated to 1 unit of our proposed grid dimension encoding.

\begin{figure}
\centering
  \includegraphics[width=1.0\linewidth]{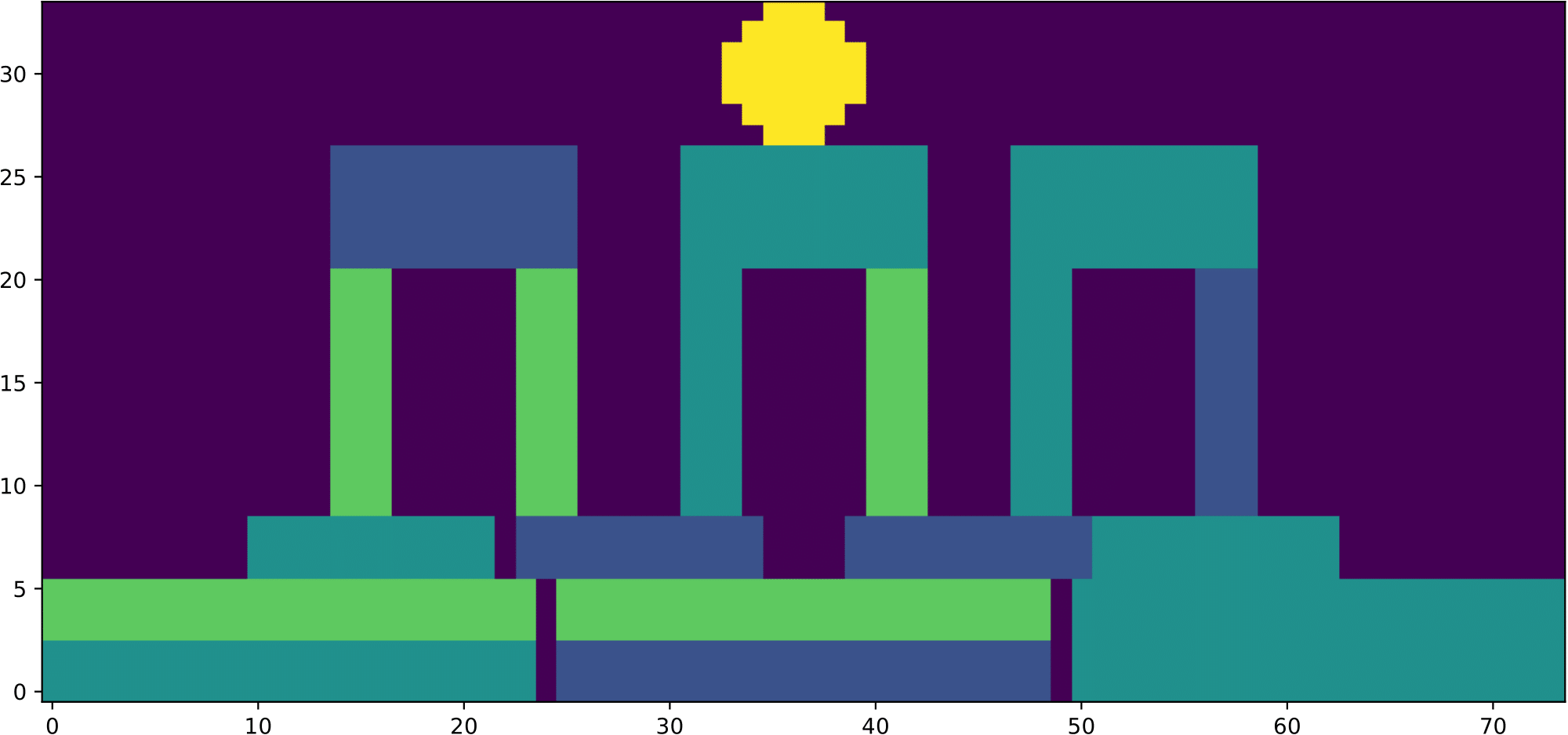}
  \caption{An encoded (rasterized) 2D image of an Angry Birds structure to a precision (resolution) of 0.07.}
  \label{fig:block_encoding_1}
\end{figure}

Using this chosen raster size, we can now convert any given XML level description into a matching grid-based encoding. For each block, the horizontal and vertical start and end positions are calculated by taking the given center position of the block and adding/subtracting half of the width/height respectively. To transform these positions into the grid indices, each value is divided by the raster size (0.07) and rounded to the nearest integer. Doing this for each object has the effect of converting its position and size to exactly fit our defined grid dimensions. Figure \ref{fig:block_encoding_1} shows how the example structure shown in Figure \ref{fig:wireframe} can be converted using this proposed encoding approach into a rasterized image.

One limitation with this structure encoding approach is that number and size of the grid cells is fixed, which places a bound on the maximum structure size that could be encoded. For our GAN model, we fixed our grid dimension at 128x128 cells or approximately 9x9 in-game distance units. This can be increased to support larger structures if desired, although this will naturally result in longer training and generation times.

\subsubsection{Multilayer Representation}

\begin{figure}
  \includegraphics[width=1.0\linewidth]{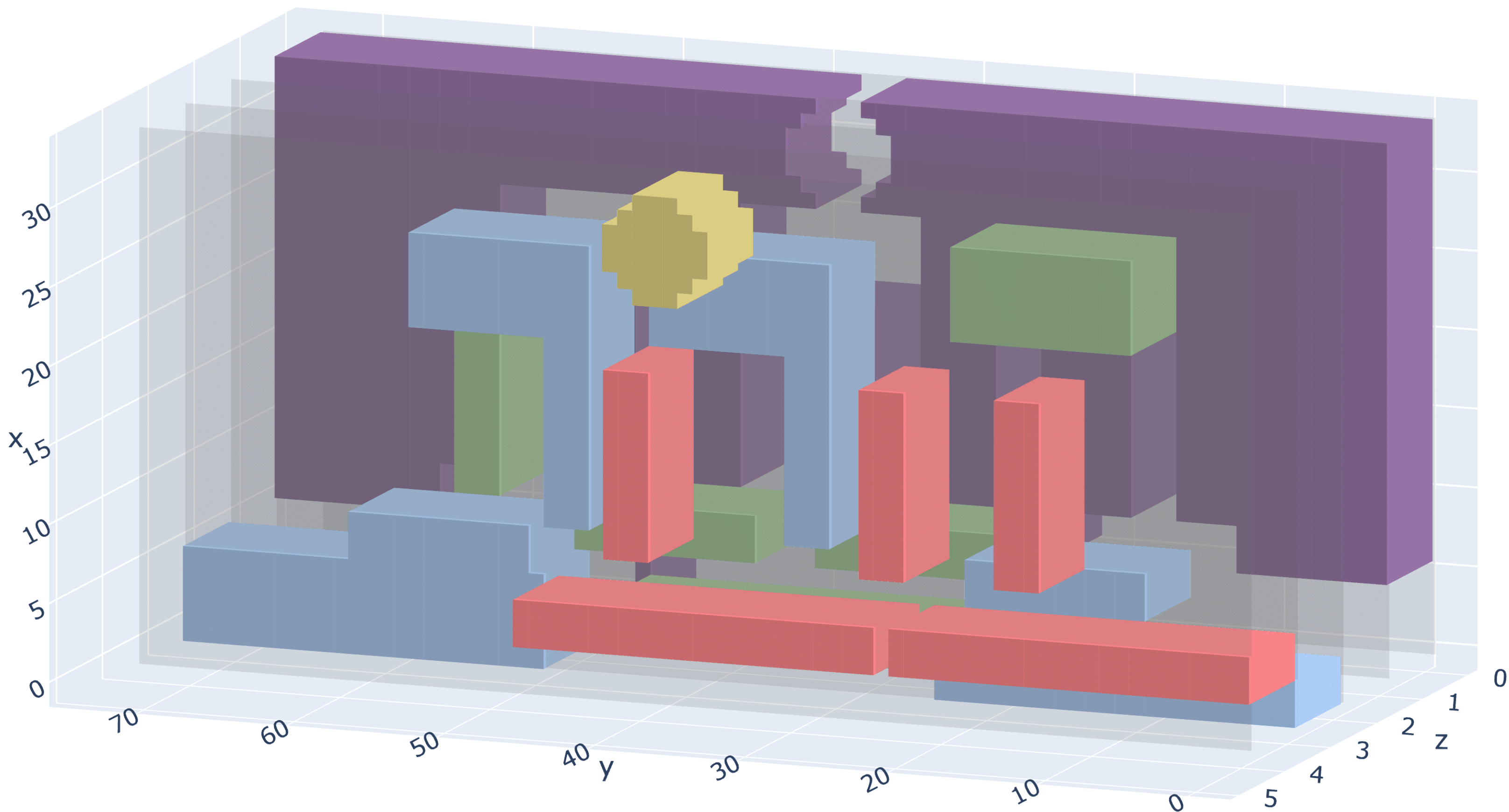}
  \caption{A multilayer representation of an Angry Birds structure, with separate layers for each object type visualized along the z-axis}
  \label{fig:air_layer}
\end{figure}

While it is possible to represent all four object types (wood/ice/stone blocks and pigs) as different values (wood = 1, ice = 2, etc.) within a single 2D grid representation, this makes training our GAN model more difficult. With this representation approach, our GAN model must simultaneously decide about the positioning of the elements and their material within a single layer.
Looking at prior work, GAN data representations are usually not only two-dimensional. For example, face
synthesis requires three channels for the RGB-Color space, and Volz et al. \cite{Volz2018}
used ten channels in a one-hot encoding, one for each Mario block type.
By moving each object type to separate layers, the decision for each layer is simplified to only predicting if an element of the associated type is present. Figure \ref{fig:air_layer} shows a visual representation of this multilayer encoding approach for the same 2D image shown in Figure \ref{fig:block_encoding_1}. Using this multilayer representation splits the blocks of each material type into separate layers, with pigs and air (i.e., empty space) each getting their own layer as well, making for a total of five layers.

\subsection{Level Decoding}

This section describes the inverse of the previous process, that of taking a structure that is output by our trained GAN model using our specified grid-based representation and decoding it into a valid XML level description for Science Birds.

\subsubsection{Confidence Decoding}

Most GAN-based approaches for procedural content generation use an encoding method that allows for a one-to-one decoding, meaning that no extra steps are required to create the level. This can potentially lead to broken elements, such as a pipe that may be missing several tiles \cite{Volz2018} and can be solved by encoding the whole group of tiles together. The decoding process can also include a healing section, where a post-processing algorithm fixes errors, tests for playability and repairs the level accordingly.

When using the multilayered encoding defined in the previous section our GAN outputs a real-valued matrix for each object type, with each matrix having the same dimensions as our defined grid representation. A value closer to one for a given matrix entry represents a higher confidence that an object of that type should be present in the corresponding grid cell, with a value close to one for the air layer representing that there should be no object present. In other words, our output matrices represent the predicted confidence that each GAN layer has for an object of its type being present at any given grid cell. This multilayer representation can be subsequently converted back into a flat 2D image using the argmax operator, where the layer with the highest confidence value output is used. However, simply determining the object type for each grid cell will not be sufficient, as our output XML representation can only include blocks of pre-defined sizes. In other words, each block is encoded into, and must also therefore be decoded from, multiple grid cells.

\begin{figure*}
    \subfloat[Selection-Ranking matrix with no clipping]{%
      \includegraphics[width=0.5\linewidth]{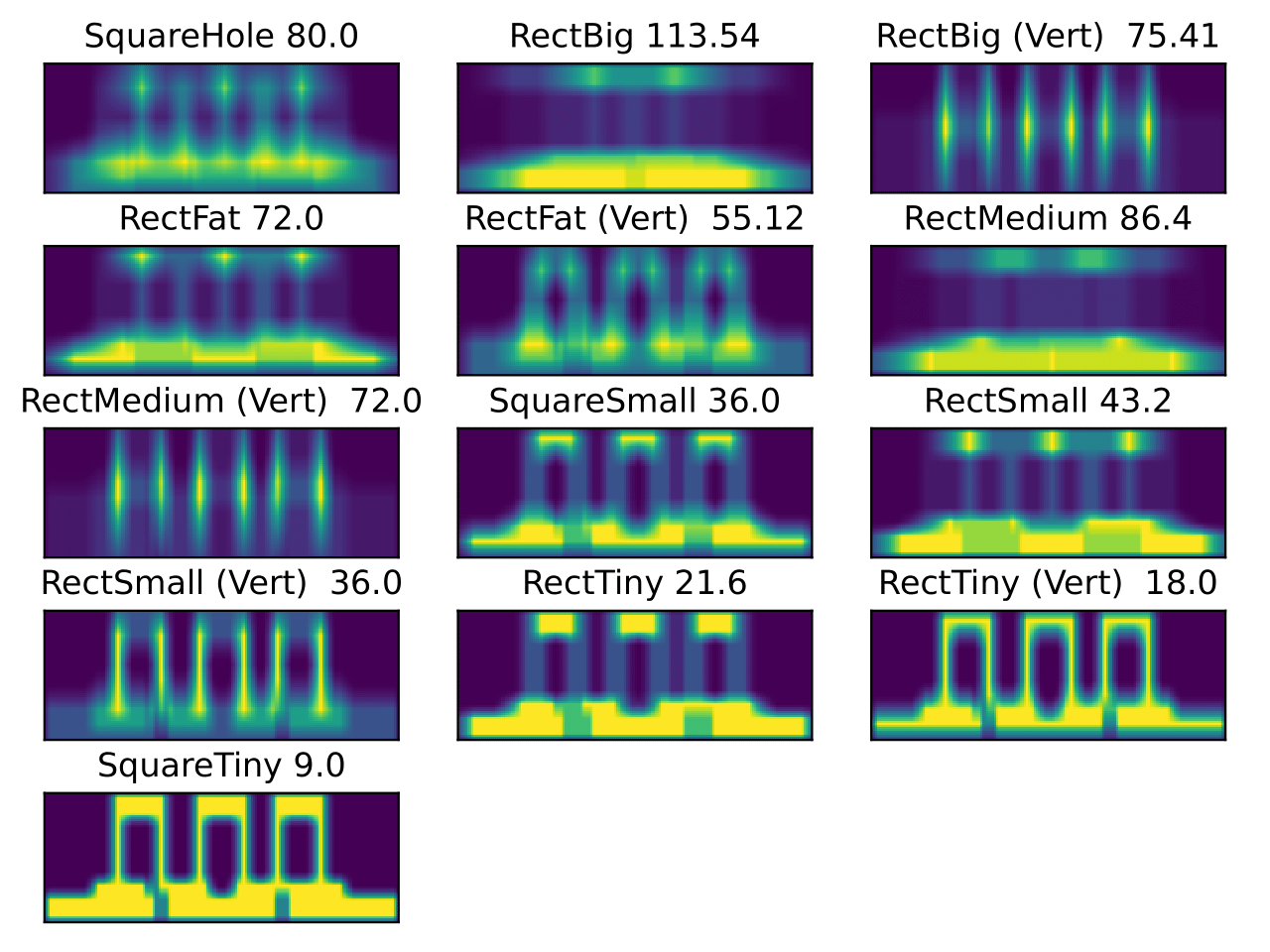}%
      \label{fig:selection_ranking}%
    }
    \subfloat[Selection-Ranking matrix with hit probabilites clipped at 0.98]{%
      \includegraphics[width=0.5\linewidth]{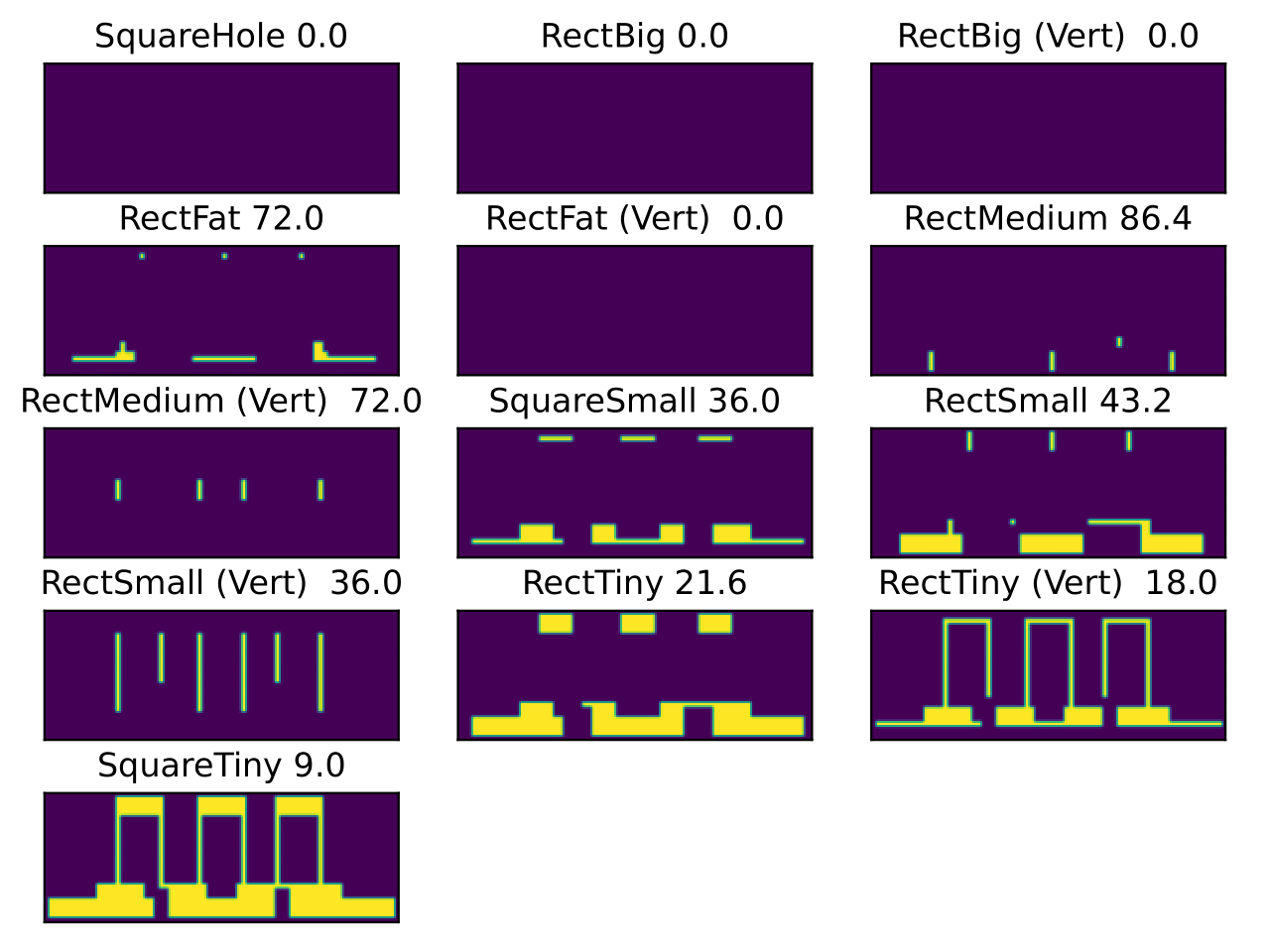}%
      \label{fig:clipped}%
    }
    \caption{
         Heatmap representations of the Selection-Ranking matrix for each block type, calculated based on the layer-wise multiplication of the Hit-Probability matrix and Size-Ranking matrix. Subfigure (a) represents the original Selection-Ranking matrix; while subfigure (b) represents the same Selection-Ranking matrix but with the Hit-Probabilities matrix clipped at 0.98. Each image represents a specific block type and corresponds to an individual layer of the Selection-Ranking matrix. The value above each image represents the maximum value present in this layer.
    }
    \label{fig:working_matirx}
\end{figure*}

\subsubsection{Matrix Creation}

To decode a generated structure image back into a valid xml level description, we need to identify suitable block positions for each material that maintain the structure's overall design. 
The first step of our proposed process involves the creation of a 3-dimensional Selection-Ranking matrix, with dimensions 128x128x13.
Each layer of this matrix represents a specific block type, including separate layers for horizontal and vertical orientations of rectangular block shapes, giving a total of 13 layers. Each of these layers contains 128x128 values representing each position of our encoded grid space, with the value at each position indicating its suitability for placing a block of the associated type. This Selection-Ranking matrix is a combination of two sub-matrices, called the Hit-Probabilities matrix and the Size-Ranking matrix.


The Hit-Probabilities matrix for each layer is created by applying a Gaussian kernel in the shape of the associated block type across the encoded structure representation, see Table \ref{fig:block_types}. Each layer of this matrix represents how well the associated block type would match the encoded structure representation if placed at each possible position (i.e., what is the percentage of overlap between the block and the structure). However, using this matrix alone leads to an issue, namely that smaller blocks (shown in the lower rows) will have a much higher hit probability than larger blocks, simply due to their smaller size. If only the highest hit probabilities were used in the selection process, any small imperfections in the representation would dismiss any larger block types that may have been intended to be there, and a group of smaller blocks would likely be used instead. 
To address this, we introduced an additional Size-Ranking matrix that applies a sum kernel to add up the values of all covered pixels for each block type at each location. 

Using the encoded Angry Birds structure representation from Figure \ref{fig:block_encoding_1} as our input, Figure \ref{fig:selection_ranking} shows the layer-wise multiplication of these two sub-matrices into the combined Selection-Ranking matrix, where yellow areas represent locations where placing the associated block type would most match the encoded structure representation.
It can be seen in Figure \ref{fig:selection_ranking} that the horizontal
RectBig layer has the highest selection value (113.54), even though it crosses gaps in the encoded structure
image. By clipping our Hit-Probabilities matrix at a high value, in this case 0.98, any block types that cross gaps such as these are removed. In practice, this has the effect of making our Selection-Ranking matrix more distinct with harder edges, see Figure \ref{fig:clipped}, and ensures that any small gaps in the original structure encoding are more likely to be maintained in the decoded output. This final clipped version of the Selection-Ranking matrix is then used as input for the subsequent block selection algorithm.

\begin{figure*}[!h]
    \centering
    \subfloat[Iteration 1]{%
      \includegraphics[width=0.47\linewidth]{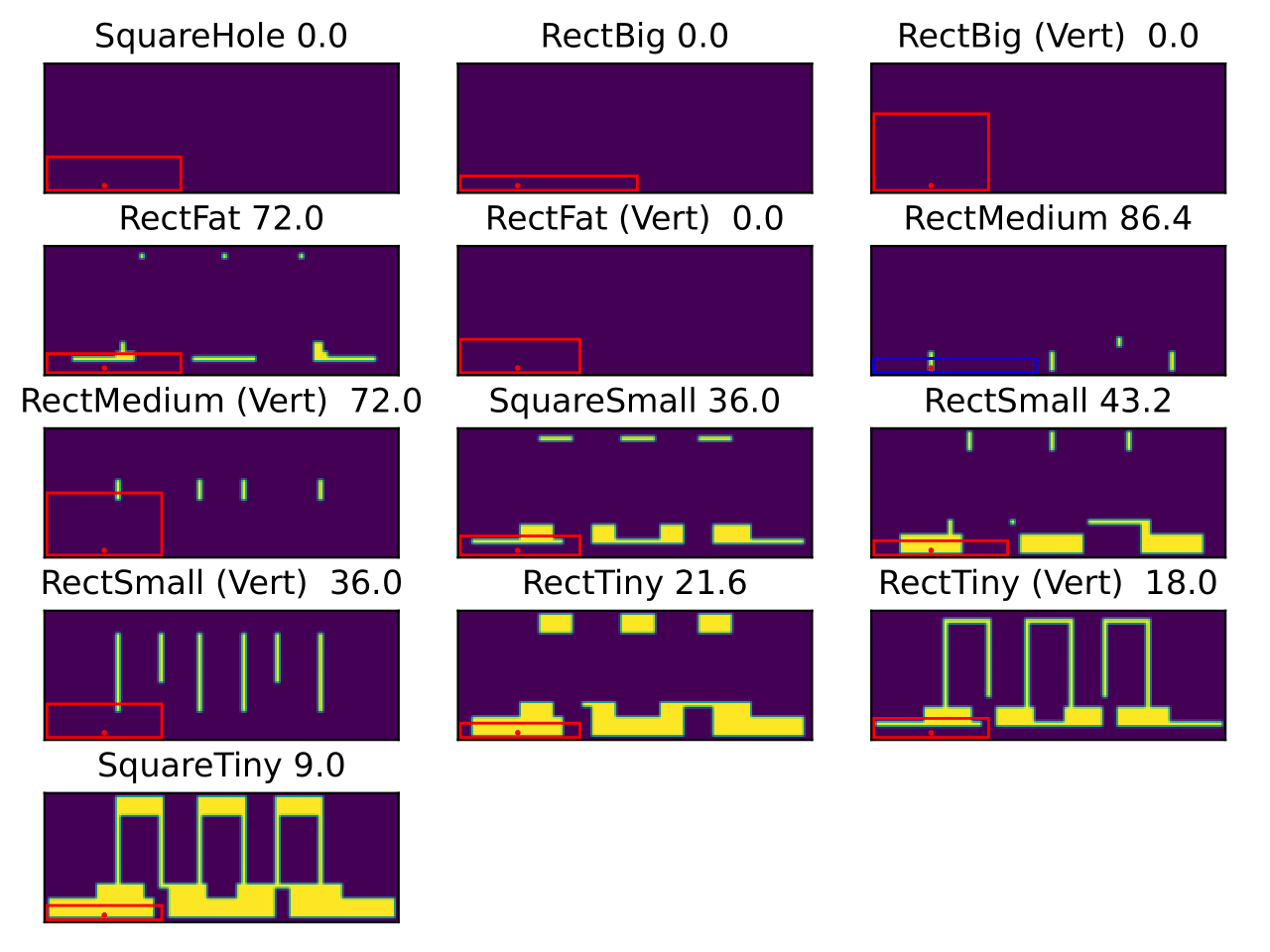}%
      \label{fig:iteration1}%
    }
    \hfill
    \subfloat[Iteration 2]{%
      \includegraphics[width=0.47\linewidth]{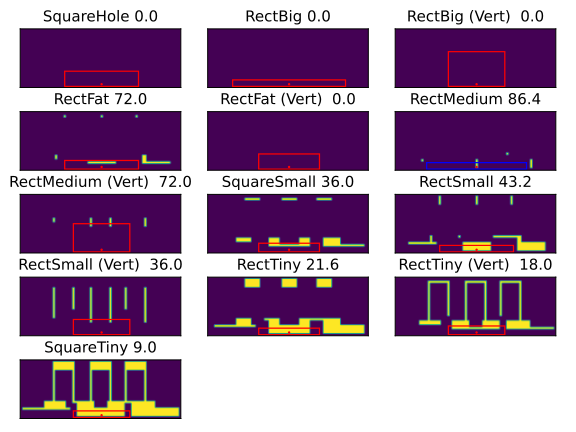}%
      \label{fig:iteration2}%
    }
    \newline
    \subfloat[Iteration 8]{%
      \includegraphics[width=0.47\linewidth]{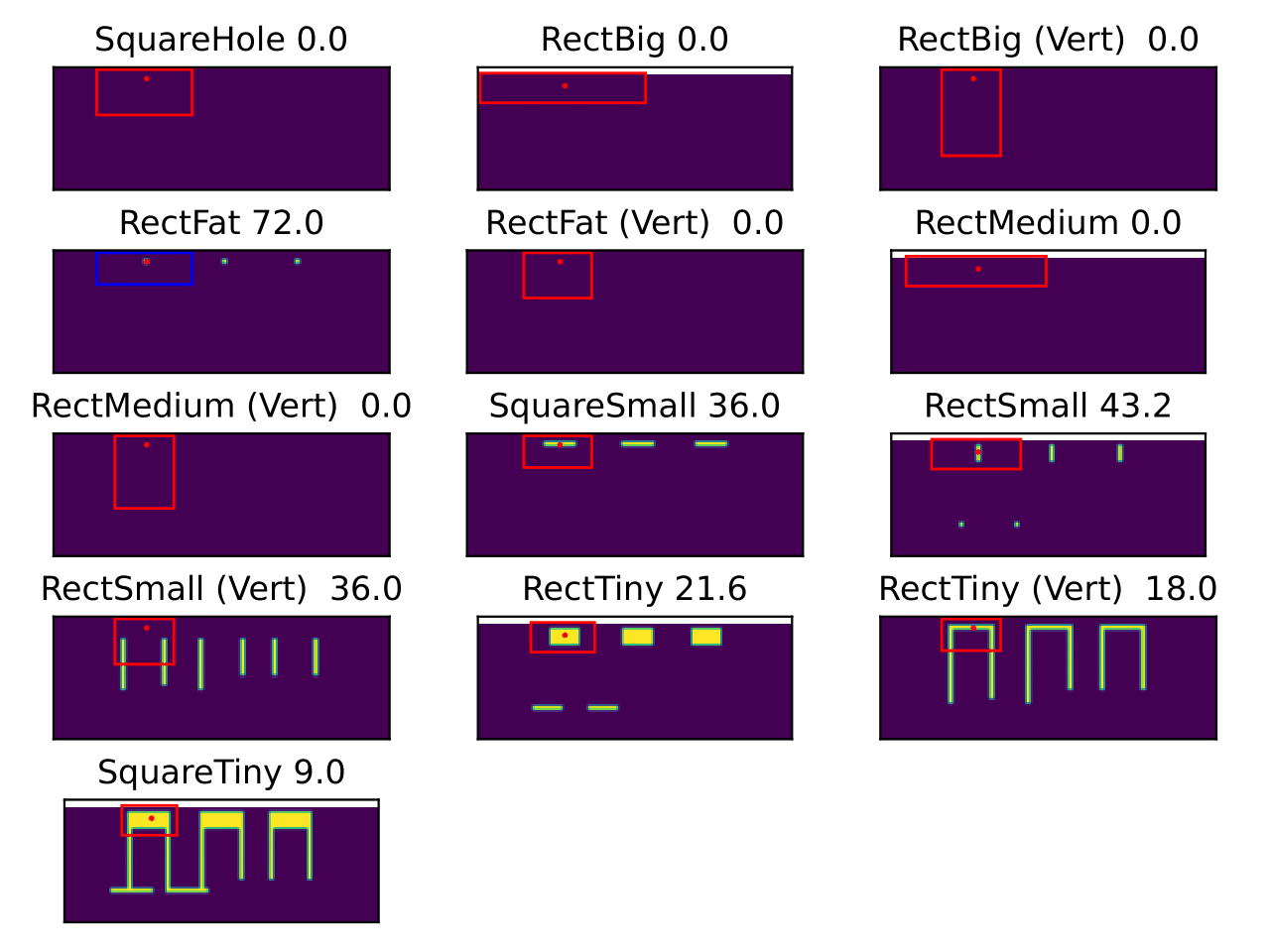}%
      \label{fig:iteration8}%
    }
    \hfill
    \subfloat[Iteration 17]{%
      \includegraphics[width=0.47\linewidth]{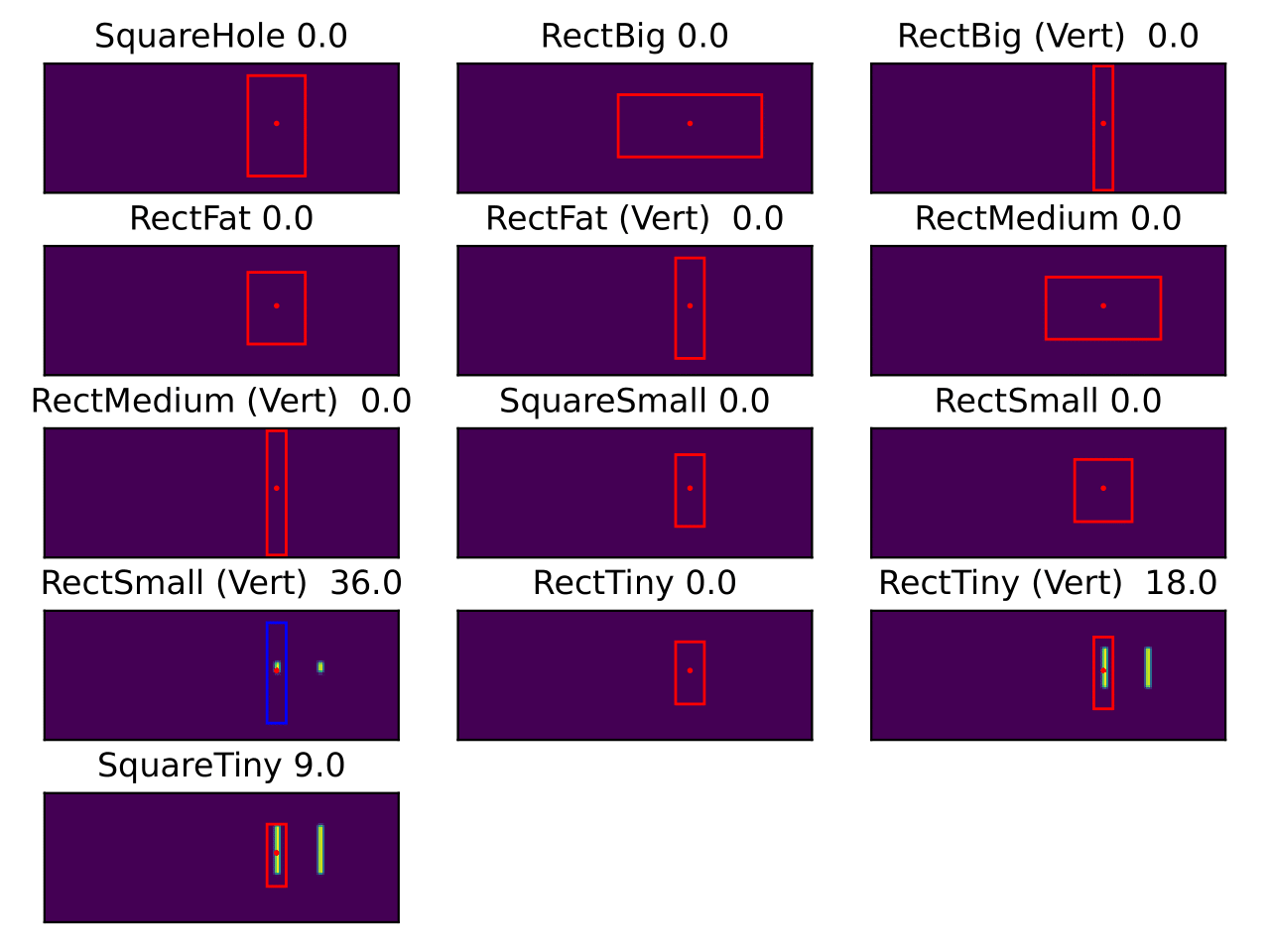}%
      \label{fig:iteration17}%
    }
    \caption{
         Heatmap representations of the clipped Selection-Ranking matrix for iterations 1, 2, 8 and 17 of the block selection process. The block type (matrix layer) and location with the highest Selection-Ranking matrix value is chosen at the start of each iteration (represented by the blue block outlines). Any values/positions in the Selection-Ranking matrix that are no longer valid, due to overlapping this selected block, are set to zero for future iterations (represented by the red outlines). This process repeats until all values in the Selection-Ranking matrix are equal to zero.
    }
    \label{fig:block_selection_iterations}
\end{figure*}


\begin{figure}[!h]
  \includegraphics[width=1.0\linewidth]{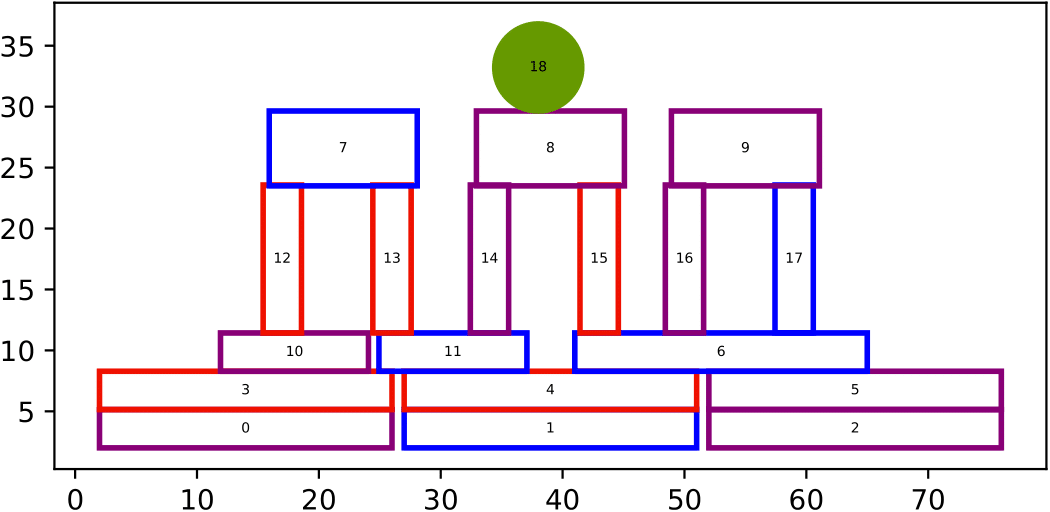}
  \caption{The result of the decoding process.}
  \label{fig:result}
\end{figure}

\subsubsection{Block Selection}

Our block selection algorithm starts by identifying the highest value in our clipped Selection-Ranking matrix, which corresponds to the ``best fit'' location for a specific block type, see blue block outline for the horizontal RectMedium layer in Figure \ref{fig:iteration1}. We then place a block of this type at this location and set any values in our matrix that would overlap this block to zero, see red outlines for the other block types/layers in Figure \ref{fig:iteration1}. This process repeats until all values in our matrix are set to zero. The clipped Selection-Ranking matrix shown in Figure \ref{fig:clipped} takes a total of 18 iterations for this process to finish, where in each iteration a new block is selected and added to the final structure output. Figure \ref{fig:block_selection_iterations} shows several iterations of the Selection-Ranking matrix throughout this block selection process. We then use a circular kernel to identify possible pig positions, followed by the same location selection process.

The last step of the decoding process is to address the issue of blocks slightly overlapping each other, which can occur when a block has an uneven width/height. We therefore apply a small structure adjustment that moves each block up and to the right, until it no longer overlaps any blocks that are below it or on its left side.

The finished result of this decoding process can be seen in Figure \ref{fig:result}. By comparing this to the original structure representation in Figure \ref{fig:wireframe} we can see that there are some minor differences between them, but the overall shape and design of the structure is sufficiently similar. This example demonstrates that our proposed encoding and decoding process is able to successfully convert between playable XML level descriptions, and grid-based structure representations that are appropriate for GAN training.

\subsection{GAN Training}

With the encoding and decoding process now defined, this section describes how our encoded structure representation is processed and used for training our GAN model.
In order to train our GAN model, two aspects need to be selected. Firstly, the specific architecture that will be used for the generator and discriminator networks. Secondly, the objective function that will be used to update the weights in these networks towards a desirable output.

\begin{figure*}
  \includegraphics[width=1.0\linewidth]{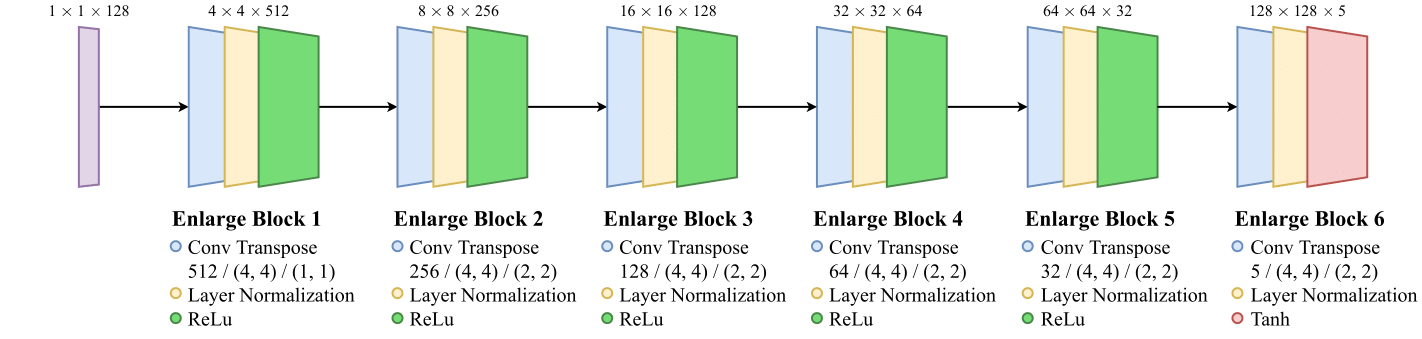}
  \caption{Generator network architecture.}
  \label{fig:GAN_generator}
\end{figure*}

\begin{figure*}
  \includegraphics[width=1.0\linewidth]{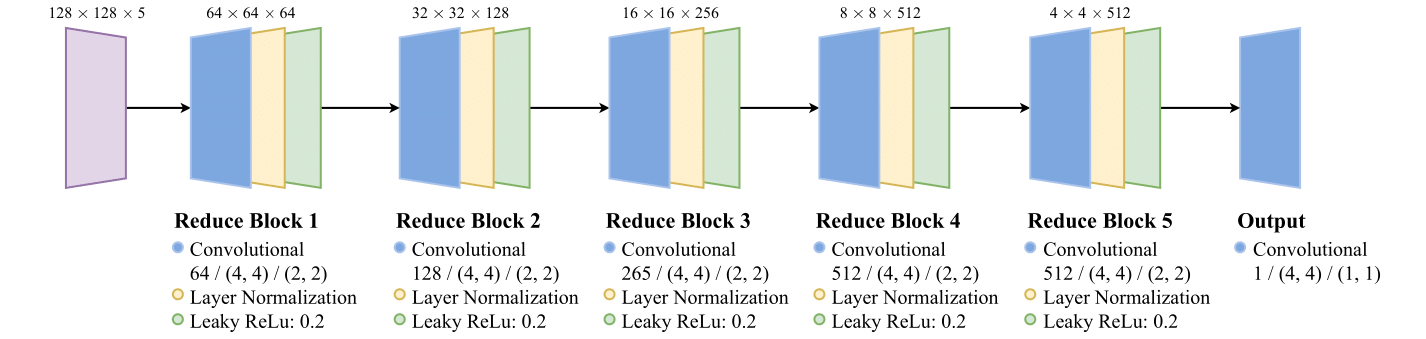}
  \caption{Discriminator network architecture.}
  \label{fig:GAN_discriminator}
\end{figure*}

\subsubsection{GAN Architecture}

Our proposed GAN model is based on the Deep Convolutional Generative Adversarial Network (DCGAN) architecture \cite{Radford2015}. This is the first Convolutional Neural Network based GAN architecture and has been shown to perform well in image generation tasks \cite{Jabbar2020}. Compared to a more standard GAN architecture, which only used fully connected layers and pooling layers, the proposed DCGAN architecture instead uses transposed convolutions to create the required image size. 

The generator network portion of our proposed GAN is shown in Figure \ref{fig:GAN_generator}. This network exclusively uses transposed convolutional layers to enlarge the
image, with a stride value of two effectively doubling the resolution size with each layer block, as this has been shown to significantly improve the stability of GAN training \cite{Jabbar2020}. The standard batch normalization layers were instead replaced by layer normalization \cite{Ba2016}, to reduce the risk of mode collapse during training. The generator uses the ReLU activation function \cite{Nair2010} between
the layers and a Tanh function at the output layer, similar to the original implementation by Goodfellow et al. \cite{Goodfellow2014}.

The discriminator network is visualized in Figure \ref{fig:GAN_discriminator}. While the generator uses transposed convolutional layers, the discriminator uses convolutional layers to reduce the image size and arrive at a decision. The discriminator uses a Leaky ReLU activation function \cite{Xu2015}, based on the recommendation by Radford et al. in their architecture guidelines for stable Deep Convolutional GANs \cite{Radford2015}.

\subsubsection{Objective Function}

For our objective function, we decided to use the Wasserstein objective function that is based on the Wasserstein or Earth Mover's distance metric \cite{Arjovsky2017Wgan}. This approach provides several benefits, including increased training stability and reduced risk of mode collapse. In addition, we use an updated version of this objective function that applies a gradient penalty to the discriminator network \cite{Gulrajani2017}, rather than the regular weight clipping approach. This change has been shown to further increase GAN training stability and convergence likelihood.

\begin{figure*}[!ht]
    \begin{minipage}{.5\textwidth}
    \raggedright
      {\includegraphics[width=0.49\linewidth]{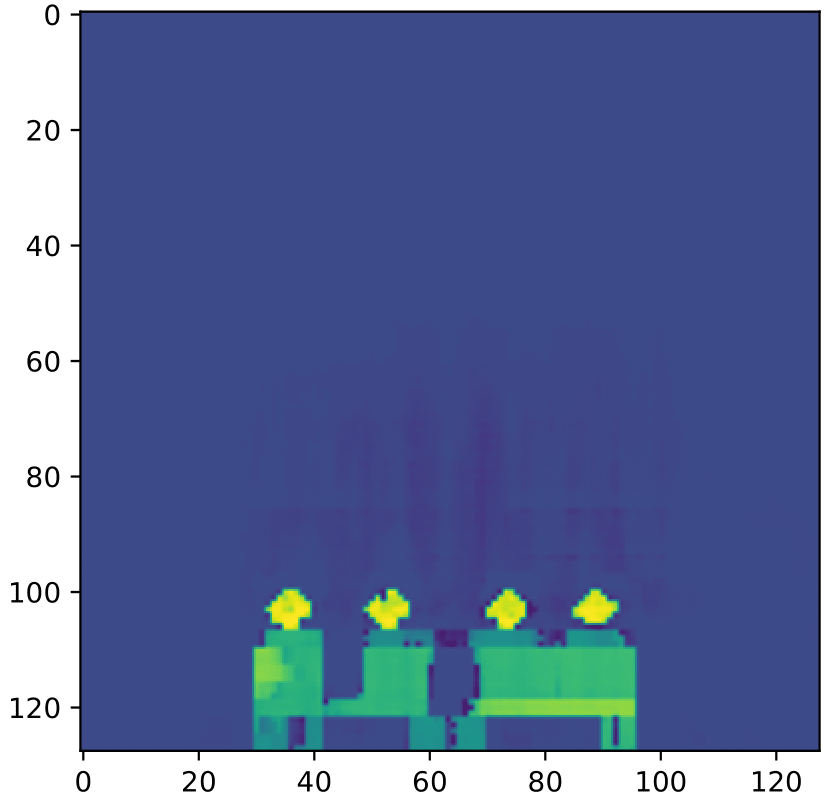}}
      {\includegraphics[width=0.49\linewidth]{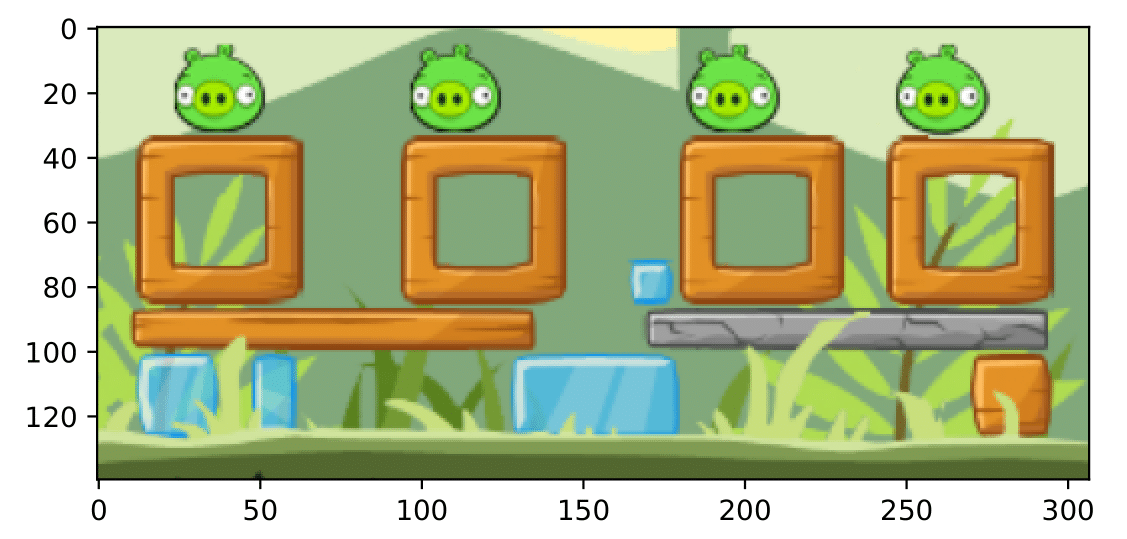}}
      {\includegraphics[width=0.49\linewidth]{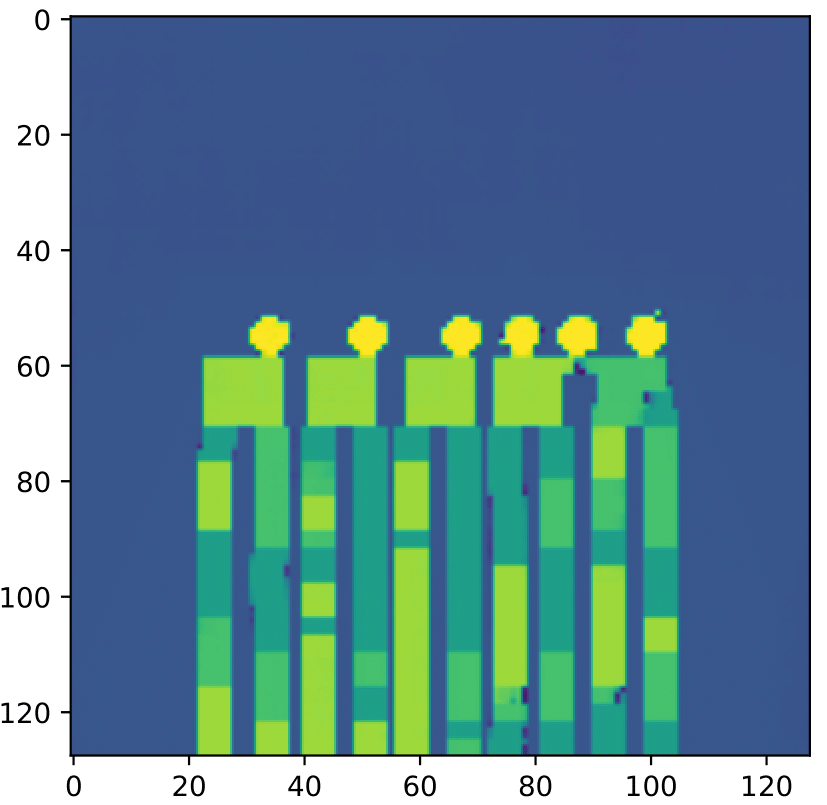}}
      {\includegraphics[width=0.49\linewidth]{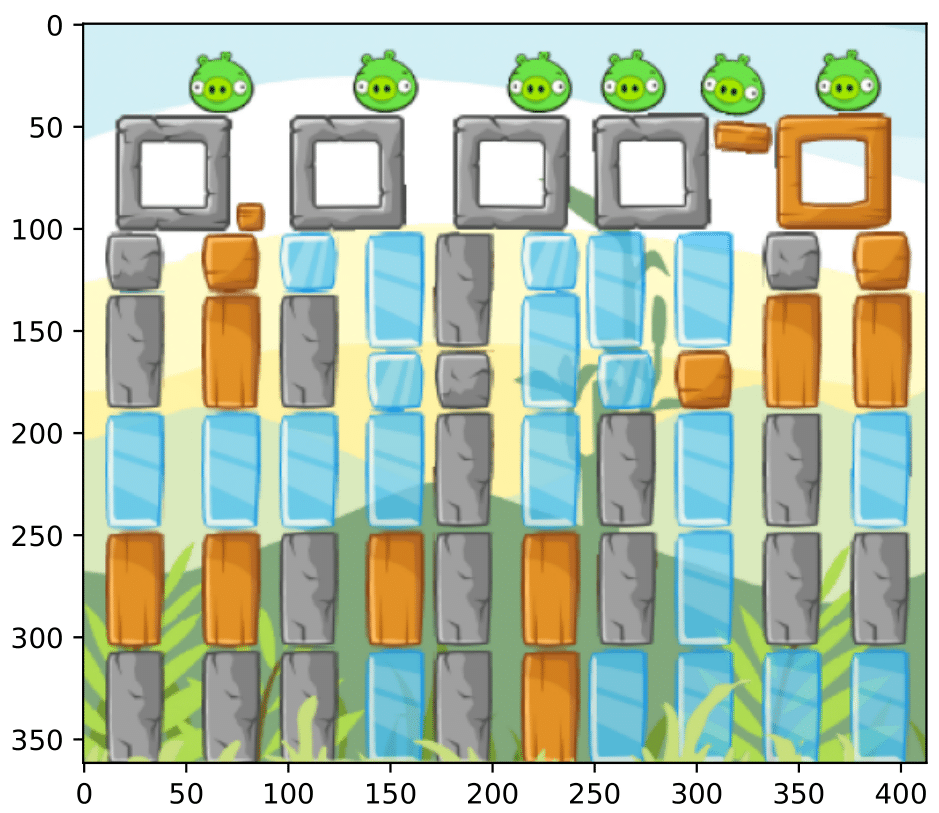}}
      {\includegraphics[width=0.49\linewidth]{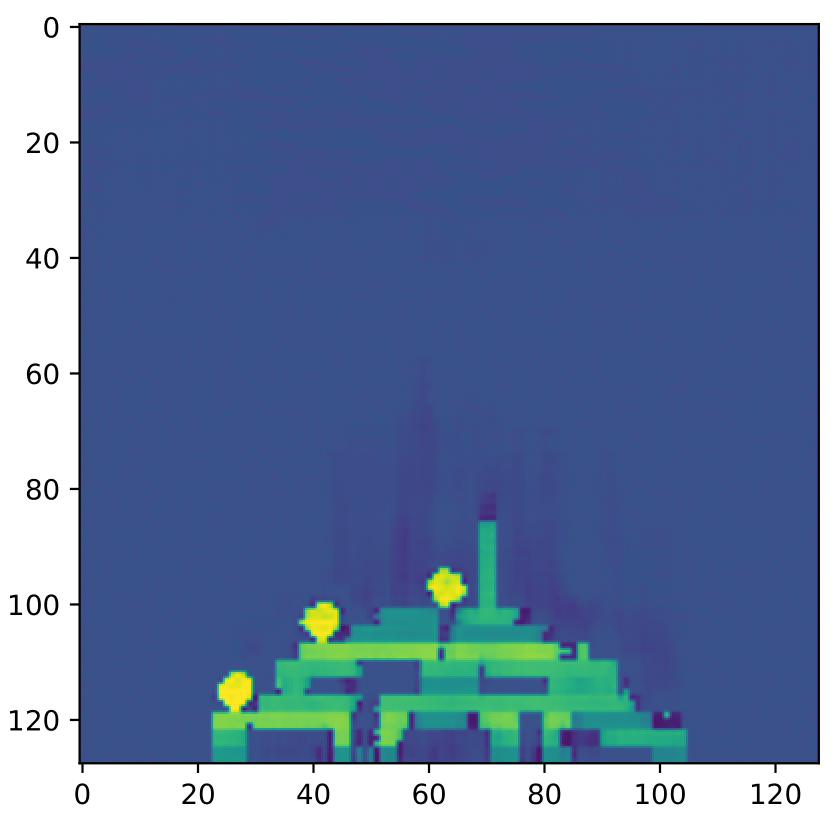}}
      {\includegraphics[width=0.49\linewidth]{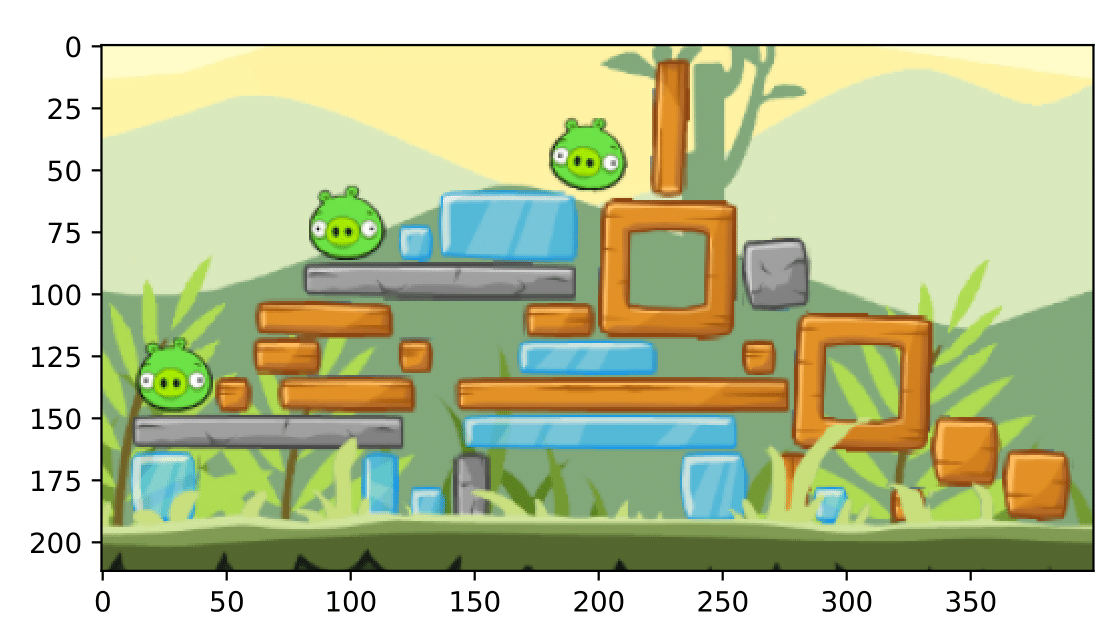}}
    \end{minipage}%
    \begin{minipage}{.5\textwidth}
    \raggedright
      {\includegraphics[width=0.49\linewidth]{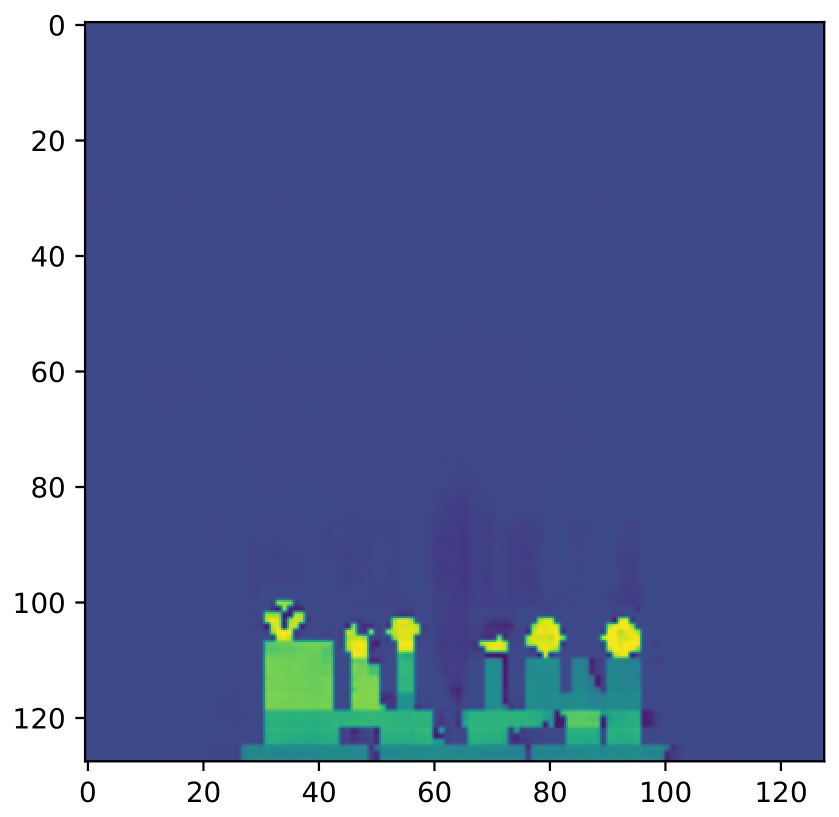}}
      {\includegraphics[width=0.49\linewidth]{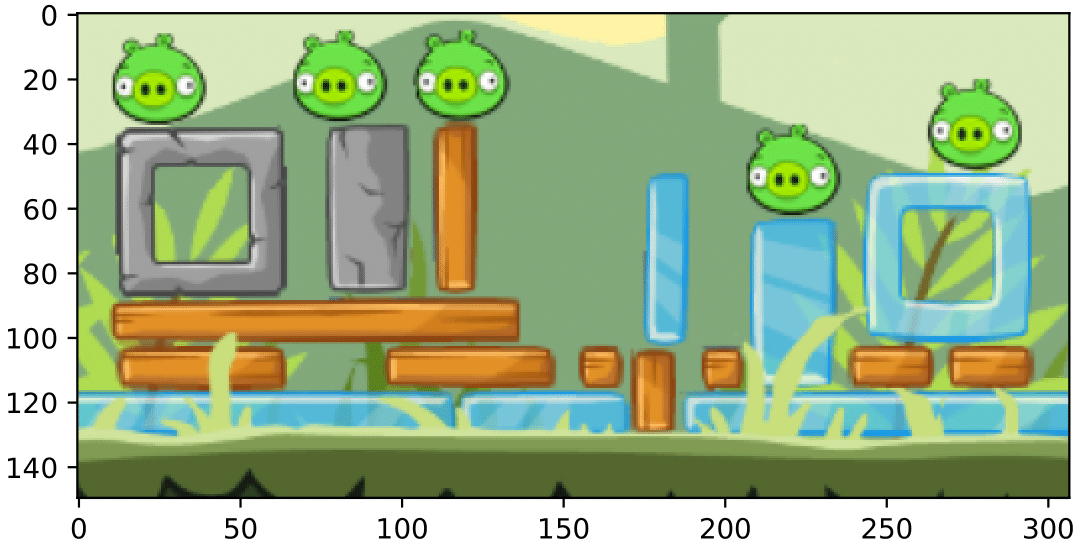}}
      {\includegraphics[width=0.49\linewidth]{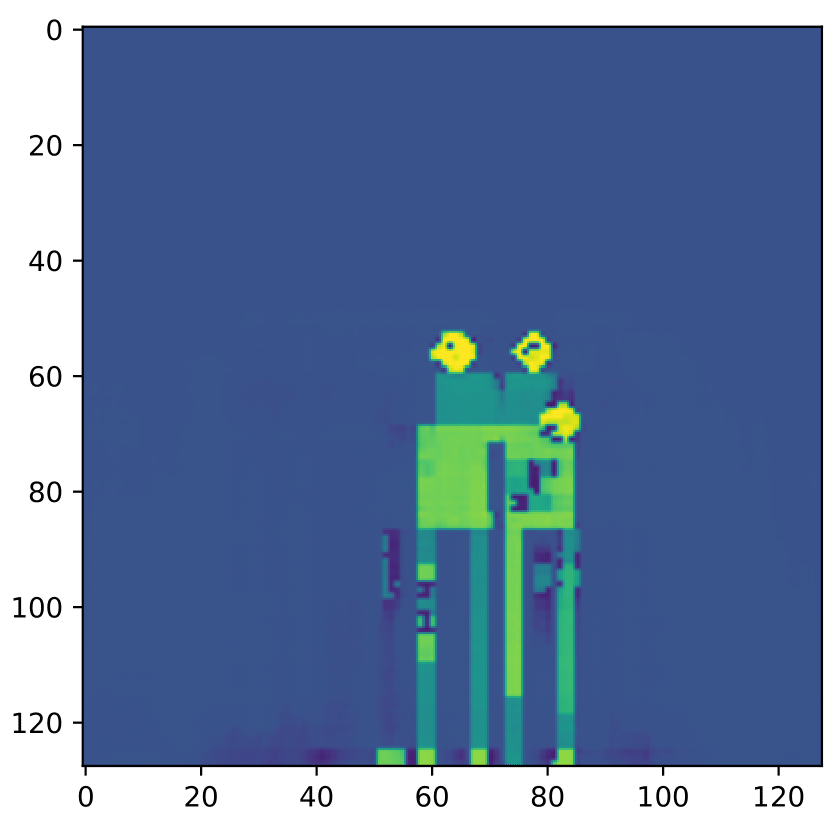}}
      {\includegraphics[width=0.25\linewidth]{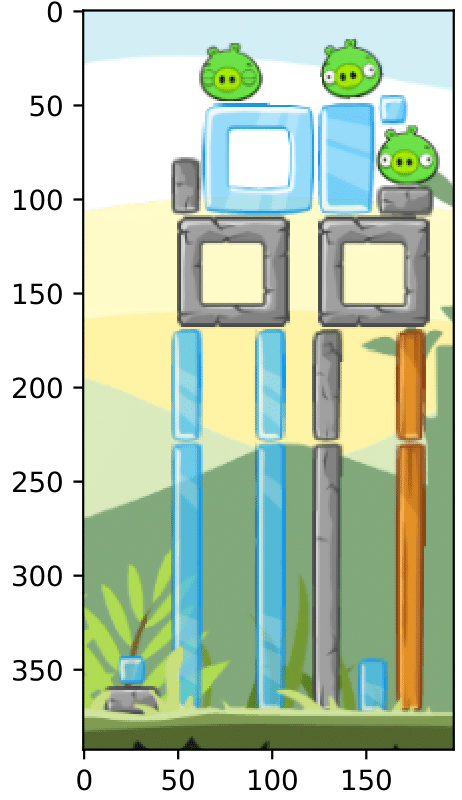}}
      {\includegraphics[width=0.49\linewidth]{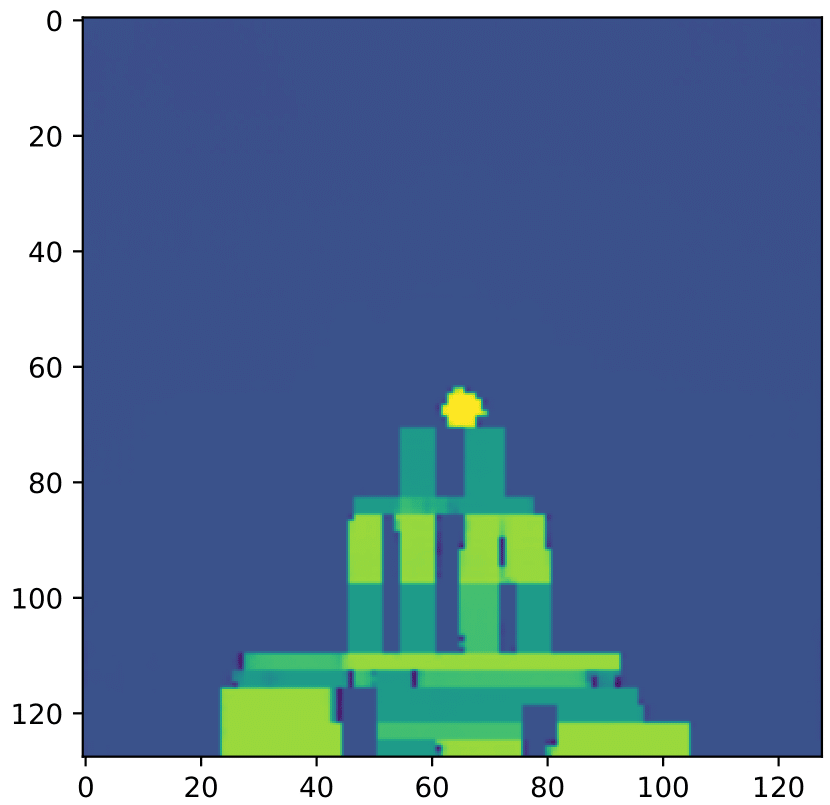}}
      {\includegraphics[width=0.49\linewidth]{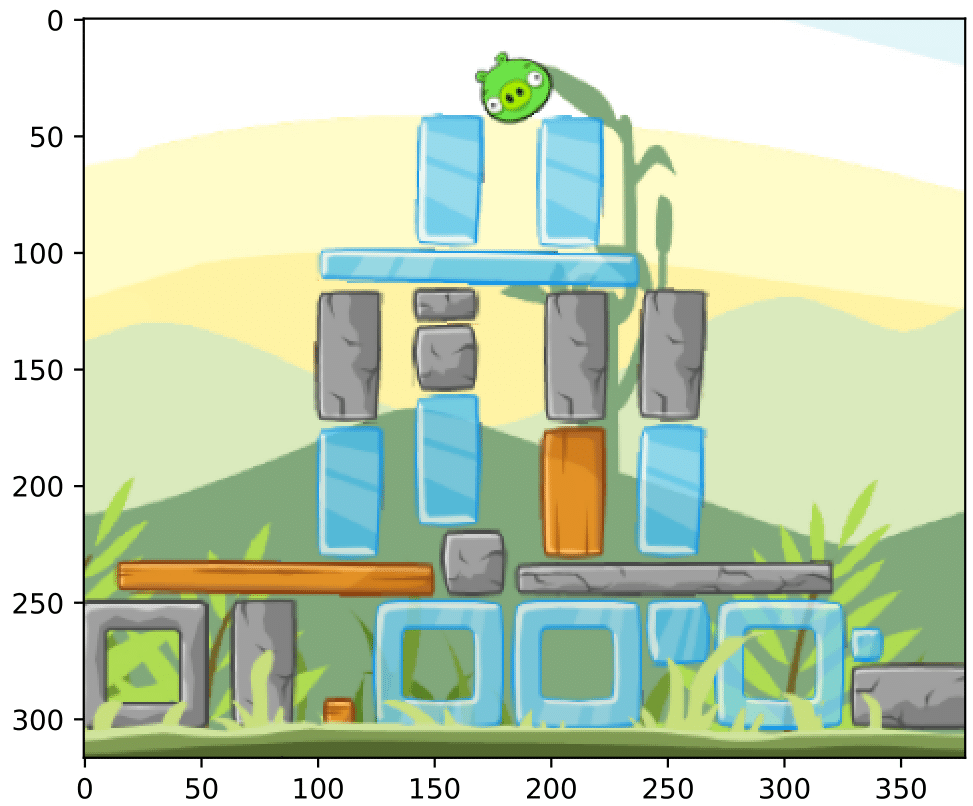}}
    \end{minipage}
    \caption{Example GAN outputs and corresponding decoded Angry Birds structures.}
    \label{fig:outputs}
\end{figure*}

\section{Experiments}

Our experiments consisted of training the proposed GAN model on an example set of encoded levels descriptions, as well as evaluating the quality and variety of output structures that it was able to produce. The complete set of all generated levels, as well as our fully trained GAN model, is open access and can be downloaded from a public GitHub repository\footnote{https://github.com/Blaxzter/Utilizing-Generative-Adversarial-Networks-for-Stable-Structure-Generation-in-Angry-Birds}.

\subsection{Model Training}

Due to the limited number of human-made levels that are currently available for Science Birds, we decided to use the open-source level generator Iratus Aves \cite{Stephenson2017} to provide structures for our training dataset. Using this generator, we created an initial training dataset of 5000 XML level descriptions that each define a single structure (including pigs).

One important consideration when training GANs is to ensure that the training dataset is sufficiently balanced in terms of its content diversity, as this can reduce the risk of mode collapse occurring. To achieve this, we applied a filter to remove any overly similar structures from our training dataset, which in turn would increase the average structure diversity. We first applied a metadata filter to remove any structures with the same number of blocks for each material, as well as the same width and height with a 0.1 unit margin. We next applied a shape filter to remove any structures with the same encoding outline, regardless of block shape or material type. After applying both these filters, our training dataset size was reduced from 5000 to 3566.

Our GAN model was trained for 15000 epochs on our filtered training dataset of 3566 generated structures. This was performed using the RWTH High-Performance Computing
cluster \cite{rwth} and took approximately 48 hours to complete.

\subsection{Results}

Our trained GAN model was used to generate 8000 different structure representations, which were then decoded into playable XML level descriptions. This process took around 15 minutes on a desktop computer with an Intel i7-5820k CPU, with over 95\% of that time being used for structure decoding. Figure \ref{fig:outputs} displays several of these GAN generated structure representation outputs, along with their decoded XML level descriptions loaded into Science Birds.
Based on these examples, we can see that many of the GAN outputs contained small distortions or noise that makes the type or shape of certain blocks ambiguous. Despite this, our proposed decoding approach can typically handle these minor imperfections and is able to produce reasonably accurate and complete structures for Angry Birds.

\subsubsection{Structure Stability}

The first test we conducted was to investigate how many of our 8000 generated structures were stable when loaded into the Science Birds game engine. The stability of any given structure can be determined using one of two approaches. The ``Block Velocity'' measure determines that a structure is stable if all blocks are stationary when the level is loaded. The ``Block Destruction'' measure determines that a structure is stable if no blocks are destroyed after the level is loaded. Blocks in Angry Birds will typically be destroyed if they fall from a sufficient height or collide with other blocks, meaning that this measure of stability is a good test for if a structure has collapsed. 

Using the Block Velocity measure, 945 structures would be classified as stable and 7055 as unstable. Using the Block Destruction measure, 3487 structures would be classified as stable and 4533 as unstable. Please note, that the Block Destruction measure is a strictly weaker version of the Block Velocity measure (i.e., any structure which is classified as stable by the Block Velocity measure is also always classified as stable by the Block Destruction measure). This large disparity would indicate that, while our generated structures often contain blocks that move slightly after loading, they are much less likely to collapse completely.

\subsubsection{Structure Diversity}

With regards to structure diversity, our generated structures varied significantly in terms of their width, height, density, shape and block frequency. Across all 8000 generated structures, the average width was 4.87 (±0.87) and the average height was 3.67 (±1.36). The average density of each structure, calculated as the percentage of the available level space that is occupied by an object, was 37.05\% (±10.56\%). The average number of blocks was 24.01 (±10.18), with an exact break down by block type and orientation provided in Table \ref{fig:block_types_frequency}. For reference, please refer to Table \ref{fig:block_types} for the dimensions of each block type. The average number of pigs in each structure was 2.76 (±1.76), with 9.1\% of generated structures containing zero pigs. While this may initially seem like a serious problem, given that a level with no pigs is already solved, it is important to remember that our proposed approach is intended to create single structures rather than complete levels. Structures without pigs can therefore still be included within a level, as long as they are placed alongside one or more structures that do contain pigs. There was also very little difference between any of these values when comparing stable and unstable structures separately.

\begin{table}[t]
\begin{tabular}{|p{0.65cm}|p{3.6cm}|p{2.8cm}|}
\hline
\textbf{Id} & \textbf{Name} & \textbf{Frequency}    \\  \hline 
1 & SquareHole & 8.75\% (±5.89\%) \\  \hline
2h & RectBig & 5.89\% (±3.20\%)   \\  \hline
2v  & RectBig (Vert) & 13.46\% (±10.21\%)    \\  \hline
3h  & RectMedium & 4.14\% (±2.13\%)  \\ \hline
3v  & RectMedium (Vert)  & 5.91\% (±4.26\%)  \\ \hline
4h  & RectSmall  & 4.18\% (±2.05\%)  \\ \hline
4v  & RectSmall (Vert)   & 7.33\% (±5.24\%)  \\ \hline
5h  & RectFat    & 5.11\% (±2.81\%)   \\ \hline
5v  & RectFat (Vert)    & 15.81\% (±16.84\%)   \\ \hline
6h  & RectTiny   & 6.84\% (±4.12\%)    \\ \hline
6v  & RectTiny (Vert)   & 6.99\% (±4.90\%)    \\ \hline
7  & SquareTiny & 9.04\% (±5.95\%)  \\ \hline
8  & SquareSmall & 6.55\% (±4.56\%) \\ \hline
\end{tabular}
\caption{Average frequency of each block type across all 8000 generated structures (±SD).}
\label{fig:block_types_frequency}
\end{table}

In terms of creating new and novel structure designs, while our generated structures appear to have several similar design elements to those present in the original training dataset, which is to be expected of a GAN based approach, they also have many differences. For example, the structures produced by the Iratus Aves generator are created using rows of blocks with the same height, leading to highly symmetrical designs \cite{Stephenson2017}. However, this was not the case for many of our GAN generated structures, such as those shown in the bottom left and top right examples for Figure \ref{fig:outputs}.

\section{Conclusion}

In this paper we have presented, implemented, trained, and evaluated a framework for using GANs to generate new structures for Angry Birds. One of the main contributions of this paper is the proposed encoding and decoding process, that can accurately convert between a playable XML level description and a grid-based structure representation more suited for GANs. Using this in conjunction with state-of-the-art GAN architectures, we were able to successfully train a GAN model to produce complete and highly varied structure designs. While many of these generated structures were initially stable, some of them unfortunately collapsed when loaded into our simulation engine. However, the number of unstable structures was not overwhelmingly large, and such structures could easily be discarded after generation via a simple generate-and-test approach. As such, we believe that the use of GAN models to generate Angry Birds structures may be used to provide an abundance of both new training content for AI agents and gameplay experiences for human players.

We would also like to highlight that this paper is a condensed version of an original Masters thesis \cite{FredericThesis}. This paper describes the most successful approach from this thesis, but several alternative encoding/decoding approaches were also investigated. We would encourage interested readers to take a closer look at this thesis for more details.

In terms of future work, one of the first extensions we might make is to train our proposed model on a larger range of content. Our experimental training set contained 3566 structures from a single generator, but there are over a dozen different level generators for Angry Birds that have been proposed during the past decade. Utilizing several of these generators could provide a much more diverse set of training levels, and potentially a more varied range of output structures. Beyond this, we could also experiment with different GAN architectures, stable diffusion models, data representations, or improvements to our encoding/decoding processes. Lastly, this approach could be applied to other physics-based domains beyond Angry Birds.

\bibliography{aaai23}

\end{document}